\documentclass[10pt,twocolumn,letterpaper]{article}

\usepackage[pagenumbers]{cvpr} 

\newcommand{\specialcell}[2][c]{%
  \begin{tabular}[#1]{@{}c@{}}#2\end{tabular}}
\usepackage{bbm}
\usepackage{multirow}
\usepackage{arydshln}
\usepackage{cancel}
\usepackage{caption}
\usepackage{wrapfig}
\usepackage{graphicx}
\usepackage{booktabs}

\definecolor{cvprblue}{rgb}{0.21,0.49,0.74}
\usepackage[pagebackref,breaklinks,colorlinks,allcolors=cvprblue]{hyperref}

\def\paperID{*****} 
\def\confName{CVPR}
\def\confYear{2025}

\title{Fractal Calibration for long-tailed object detection}

\author{Konstantinos Panagiotis Alexandridis$^{1,*}$, Ismail Elezi$^1$, Jiankang Deng$^2$, Anh Nguyen$^3$ and Shan Luo$^4$\\
$^1$Huawei Noah's Ark Lab; $^2$Imperial College London; $^3$University of Liverpool; $^4$King's College London\\
{\tt\small $^{*}$konstantinos.alexandridis@huawei.com}\\
}

\begin{document}
\maketitle
\begin{abstract}
Real-world datasets follow an imbalanced distribution, which poses significant challenges in rare-category object detection. Recent studies tackle this problem by developing re-weighting and re-sampling methods, that utilise the class frequencies of the dataset. However, these techniques focus solely on the frequency statistics and ignore the distribution of the classes in image space, missing important information.
In contrast to them, we propose \textbf{FRA}ctal \textbf{CAL}ibration (FRACAL): a novel post-calibration method for long-tailed object detection. FRACAL devises a logit adjustment method that utilises the fractal dimension to estimate how uniformly classes are distributed in image space. During inference, it uses the fractal dimension to inversely downweight the probabilities of uniformly spaced class predictions achieving balance in two axes: between frequent and rare categories, and between uniformly spaced and sparsely spaced classes. FRACAL is a post-processing method and it does not require any training, also it can be combined with many off-the-shelf models such as one-stage sigmoid detectors and two-stage instance segmentation models. FRACAL boosts the rare class performance by up to $8.6\%$ and surpasses all previous methods on LVIS dataset, while showing good generalisation to other datasets such as COCO, V3Det and OpenImages. We provide the code at \url{https://github.com/kostas1515/FRACAL}.
\end{abstract}    
\section{Introduction}
\label{sec:intro}
In recent years, there have been astonishing developments in the field of object detection 
\cite{carion2020end,chen2022diffusiondet,lyu2022rtmdet}. Most of these works utilise vast, balanced, curated datasets such as ImageNet1k \cite{deng2009imagenet}, or MS-COCO \cite{lin2014microsoft} to learn efficient image representations. 
However, in the real world, data are rarely balanced, in fact, they follow a long-tailed distribution \cite{liu2019large}. 
When models are trained with long-tailed data, they perform well for the frequent classes of the distribution but they perform inadequately for the rare classes \cite{wang2020devil,Ren2020balms,li2020overcoming}. 
This problem poses significant challenges to the safe deployment of detection and instance segmentation models in real-world safe-critical applications such as autonomous vehicles, medical applications, and industrial applications, scenarios where rare class detection is paramount. 

Many approaches address the long-tailed detection problem by employing adaptive re-weighting or data resampling techniques to handle imbalanced distributions~\cite{wang2021seesaw,wang2021adaptive,zang2021fasa}. However all these methods require training.
In contrast, in long-tailed image classification, alternative methods focus on mitigating class imbalance during inference through a post-calibrated softmax adjustment (PCSA)
~\cite{alexandridis2023inverse,Ren2020balms,hong2021disentangling}.
PCSA boasts strong performance, good compatibility with many methods like data augmentation, masked image modeling, contrastive learning, and does not necessitate specialized loss function optimization, making it more user friendly ~\cite{xu2023learning,cui2021parametric,zhu2022balanced}. 

\begin{figure}[t]
    \centering
    \includegraphics[width=1\linewidth]{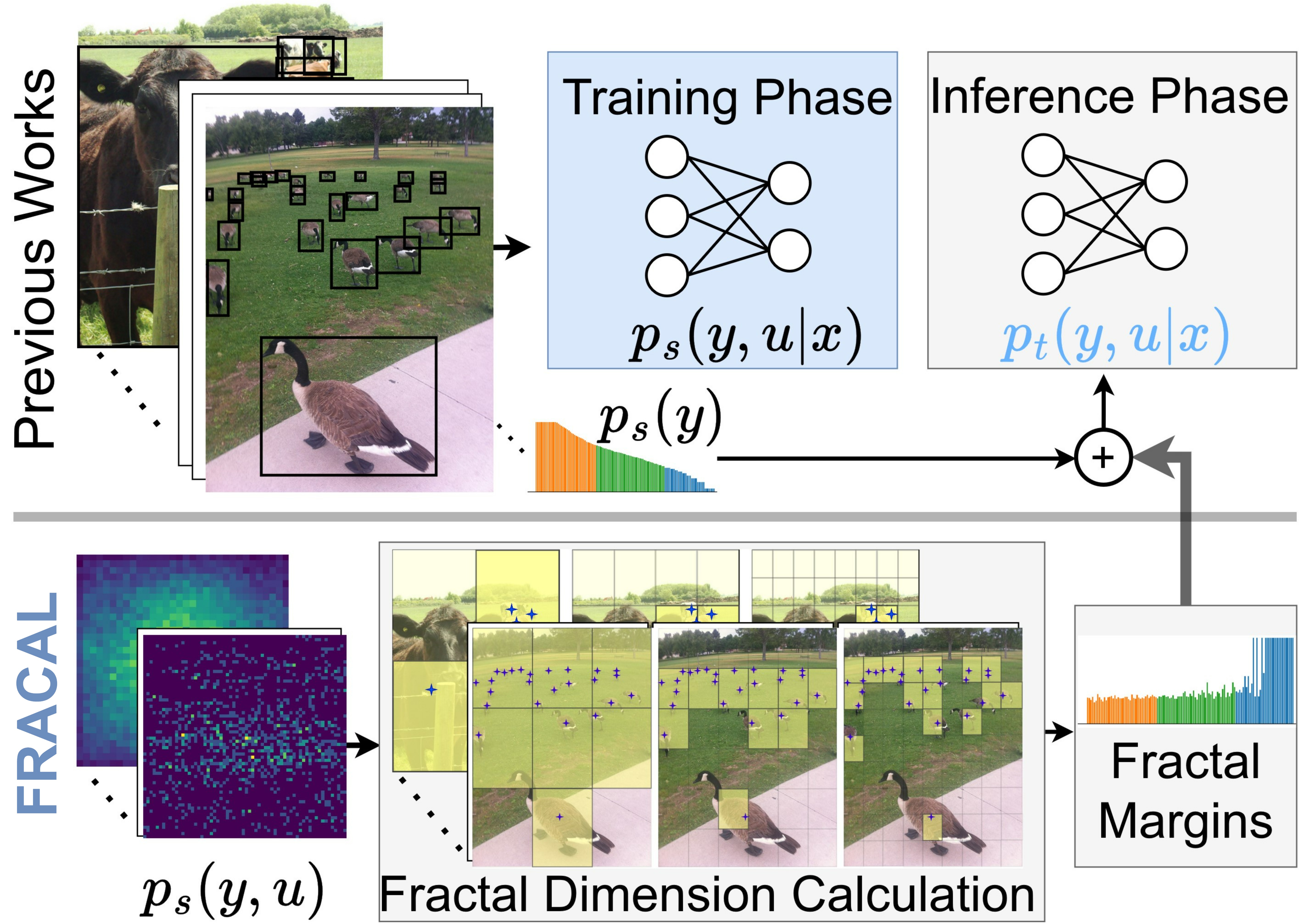}
    \caption{Previous works used class information $p_s(y)$ to align the learned source distribution $p_s(y,u|x)$ with the balanced target distribution $p_t(y,u|x)$, without considering the space $u$ and class $y$ relationship i.e. $p_s(y,u)$. FRACAL captures $p_s(y,u)$, using the fractal dimension, and embeds fractal margins during inference, aligning the learned distribution $p_s(y,u|x)$ with the target $p_t(y,u|x)$ better than previous works.}
    \label{fig:motivation}
\end{figure}
However, current PCSA methods utilise solely the train set's class frequency $p_s(y)$ as shown in Fig.\ref{fig:motivation}-top, overlooking the significance of the classes' dependence on the location distribution $p_s(y,u)$. This is a significant limitation of previous PCSA methods because the location information is a critical indicator considering the correlation between classes $y$ and their respective locations $u$.

Motivated by the class-location dependence~\cite{kayhan2020translation}, in this work, we investigate a novel way to incorporate location information into post-calibration for imbalanced object detection to boost the performance of rare classes by fully exploiting dataset statistics. 
We empirically show that naively injecting location statistics results in inferior performance because the location information is sparse for the rare classes. To overcome this, we propose FRACAL (FRActal CALibration), a novel post-calibration method based on the fractal dimension, as shown in Fig.\ref{fig:motivation}-bottom.
Our method aggregates the location distribution of all objects in the training set, using the box-counting method \cite{schroeder2009fractals}. This resolves the sparsity problem and significantly enhances the performance of both frequent and rare classes as shown in our experiments.
Our method comes with several advantages. 
First, it performs an effective class calibration, suitable for the object detection task, using the dataset's class frequencies. 
Secondly, it captures the class-location dependency~\cite{kayhan2020translation}, using the fractal dimension, and it fuses this information into class calibration. This results in a better and unique space-aware logit-adjustment technique that complements the frequency-dependent class calibration method and achieves higher overall performance compared to previous PCSA techniques.
FRACAL can be easily combined with both one-stage and two stage detectors, Softmax and Sigmoid-based models, various instance segmentation architectures, various backbones, sampling strategies, and largely increase the performance during the inference step. 
FRACAL significantly advances the performance on the challenging LVISv1 benchmark \cite{gupta2019lvis} with no training, or additional inference cost by $8.6\%$ rare mask average precision ($AP^m_r$). 

Our \textbf{contributions} are as follows:
\begin{itemize}
    \item For the first time, we show the importance of the class-location dependence in post-calibration for long-tailed object detection.
    \item We capture the location-class dependence via a space-aware long-tailed object detection calibration method based on the fractal dimension.
    \item Our method performs remarkably on various detectors and backbones, on both heavily imbalanced datasets such as LVIS and less imbalanced datasets such as COCO, V3DET and OpenImages, outperforming the state-of-the-art by up to $8.6\%$.
\end{itemize}
\section{Related Work}
\label{sec:related_work}

\begin{table}[t]
    \centering
    \caption{Post-calibration techniques in long-tailed tasks. $\tau$ and $\gamma$ are hyper-parameters, $bg$ is the background class, $\mu_y$ and $\varsigma_y$ are estimated class mean and standard deviation respectively. Compared to past works, FRACAL uses both frequency ($\mathbf{F}$) and space ($\mathbf{S}$) information, as shown in Section \ref{sec:method}.}
    \vspace*{-1em}
    \begin{tabular}{c|p{0.6cm}|p{4.75cm}}
    \hline
         Method&Use&Adjustment \\
         \hline
         LA.~\cite{menon2021longtail}&$\mathbf{F}$&$z_{y}^\prime=z_{y}-\tau\log(p_s(y))$\\
         IIF~\cite{alexandridis2023inverse} &$\mathbf{F}$&$z_{y}^\prime=-z_{y} \cdot \log(p_s(y))$\\
         PCSA~\cite{hong2021disentangling}&$\mathbf{F}$&$z_{y}^\prime=z_{y}-\log(p_s(y)) +\log(p_t(y))$\\
         Norcal~\cite{pan2021model}&$\mathbf{F}$&$p_{y}^\prime= \frac{p_{y}/n_y^\gamma}{p_{bg}+\sum p_{y}/n_y^\gamma},y\notin bg$ \\
         LogN~\cite{zhao2022logit}&$\mathbf{F}$&$z_{y}^\prime=\frac{z_{y}-(\mu_y-\min_y(\mu_y))}{\varsigma_y},y\notin bg$\\
         \midrule
         FRACAL&$\mathbf{S}$+$\mathbf{F}$&$z_{y}^\prime = \text{S}(\text{C}(z_{y})) /{\sum_{j=1}^{C+1} \text{S}(\text{C}(z_{y}))}$\\
        \hline
    \end{tabular}
    \label{tab:sim_adj_methods}
\end{table}

\noindent \textbf{General Object Detection.}
General object detection \cite{redmon2017yolo9000,ren2015faster,lin2017focal,liu2016ssd,carion2020end,zhu2021deformable,sun2021sparse,chen2022diffusiondet,li2022exploring} and instance segmentation \cite{he2017mask,huang2019mask,cai2019cascade,chen2019hybrid,wang2019carafe,bolya2019yolact,li2022exploring} have witnessed tremendous advancements.
Recently, transformer-based detectors were proposed which use self-attention to directly learn object proposals \cite{carion2020end,zhu2021deformable}, or diffusion-based methods which use a de-noising process to learn bounding boxes \cite{chen2022diffusiondet} and segmentation masks \cite{gu2022diffusioninst}. However, all of these methods struggle to learn the rare classes when trained with long-tailed data \cite{gupta2019lvis,oksuz2020imbalance} due to the insufficient rare samples. To this end, FRACAL enhances the rare class performance using a space-aware logit adjustment that can be easily applied during inference.
\begin{figure*}[t]
    \centering
    \includegraphics[width=0.9\linewidth]{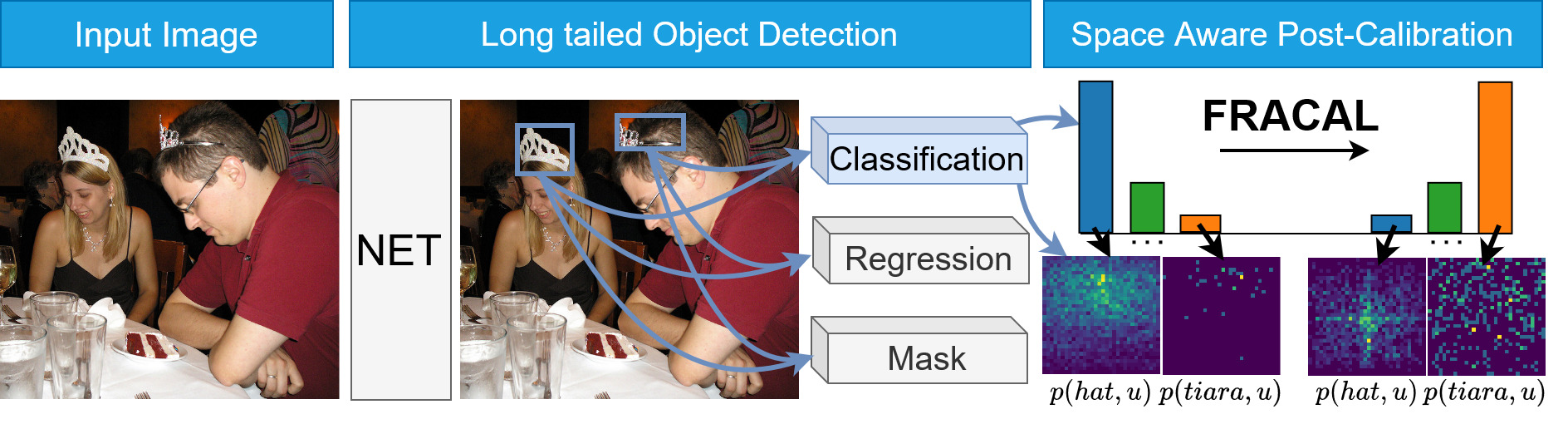}
    \caption{During imbalanced object detection, the model makes more frequent class detections like \textit{hat} and less rare class detections like \textit{tiara} both of which have strong upper location bias. FRACAL utilises fractal dimension and debiases the logits both in the frequency and space axes, making fewer \textit{hat} detections and more \textit{tiara} detections that are both evenly spread in image space.}
    \label{fig:pipeline}
\end{figure*}

\noindent \textbf{Long-tailed image classification.}
In the past years, the long-tailed image recognition problem has received great attention, as demonstrated by many recent surveys \cite{oksuz2020imbalance,zhang2023deep,yang2022survey} and newly created benchmarks \cite{yang2022multi,tang2022invariant,gu2022tackling}. 
In long-tailed classification, the works could be split into two groups, representation learning and classifier learning. Representation learning techniques aim to efficiently learn rare class features using oversampling~\cite{park2022majority,hong2022safa,zang2021fasa}, contrastive learning \cite{li2022targeted,zhu2022balanced,cui2023generalized}, using ensemble or fusion models~\cite{wang2021long,li2022nested,li2022trustworthy,cui2022reslt,aimar2023balanced}, knowledge distillation~\cite{li2022nested,he2021distilling,li2021self}, knowledge transfer~\cite{liu2019large,parisot2022long,zhu2020inflated}, sharpness aware minimisation \cite{zhou2023imbsam,zhou2023class,ma2023curvature} and neural collapse \cite{li2023fcc,zhong2023understanding,liu2023inducing}. Classifier learning techniques aim to adjust the classifier in favour of the rare classes via decoupled training~\cite{kang2019decoupling,zhang2021distribution,hsu2023abc}, margin adjustment~\cite{menon2021longtail,Ren2020balms,hong2021disentangling,cao2019learning,hyun2022long,zhao2022adaptive,alexandridis2023inverse,ye2020identifying} and cost-sensitive learning~\cite{cui2019class,khan2017cost,wang2017learning}. Among these works, the Post-Calibrated Softmax Adjustment (PCSA) method \cite{menon2021longtail,hong2021disentangling,threeHead} distinguishes itself through both its strong performance and the absence of any training requirements.
However, most of the classifier and representation learning techniques are hard to adopt in long-tailed object detection. This difficulty arises from the larger imbalance inherent in this task, amplified by the presence of the background class \cite{mullapudi2021background,yang2022survey}. Moreover, the optimisation of models for this task becomes more complex due to multiple sources of imbalance such as batch imbalance, class imbalance and task imbalance as outlined in this survey \cite{oksuz2020imbalance}. For this reason, we develop FRACAL, which is a post-calibration method tailored to the long-tailed object detection task. Different from post-calibration classification methods \cite{menon2021longtail,hong2021disentangling}, FRACAL enhances the detection performance by leveraging class-dependent space information derived from the fractal dimension. Through space-aware logit-adjustment, FRACAL mitigates biases in both the detection's location and classification axes.

\noindent \textbf{Long-tailed object detection.}
The most prevalent technique is adaptive rare class re-weighting, which could be applied using either the statistics of the mini-batch ~\cite{hsieh2021droploss,tan2020equalization,wang2021adaptive} or the statistics of the
gradient \cite{tan2021equalization,li2022equalized}. Other works use adaptive classification margins based on the classifier's weight norms~\cite{wang2022c2am,li2022adaptive}, classification score~\cite{feng2021exploring,he2022relieving,wang2021seesaw}, activation functions \cite{alexandridis2022long,alexandridis2024adaptive}, group hierarchies \cite{li2020overcoming,wu2020forest} and ranking loss \cite{zhang2023reconciling}. Many works use data resampling techniques \cite{zang2021fasa,gupta2019lvis,kang2019decoupling,feng2021exploring,wu2020forest} or external rare class augmentation \cite{zhang2022learning,zhang2021mosaicos}. 
All these works optimise the model on the long-tailed distribution and require the construction of a complicated and cumbersome training pipeline. In contrast, our method operates during the model's inference stage thus it is easier to use and less evasive to the user's codebase.

Norcal~\cite{pan2021model} was the first method to apply a post-calibration technique in imbalanced object detection, achieving promising results without training the detector. They proposed to calibrate only the foreground logits using the train-set's label statistics and applied a re-normalisation step. LogN \cite{zhao2022logit} proposed to use the model's own predictions to estimate the class statistics and applied standardisation in the classification layer.

However LogN, requires forward-passing the whole training set through the model to estimate the weights, thus it is slower than FRACAL, which is not model-dependent.
Also, both methods do not utilise the spatial statistics of the classes which are valuable indicators since the classes and their location are correlated \cite{kayhan2020translation}. 

To this end, FRACAL balances the detectors using both class and space information, largely surpassing the performance of the previous methods. FRACAL can be easily combined with two-stage softmax-based models like MaskRCNN \cite{he2017mask}, or one-stage sigmoid detectors such as GFLv2~\cite{li2021generalized} achieving great results without training or additional inference cost.

\noindent \textbf{Relation to previous works.}
In Table \ref{tab:sim_adj_methods}, we contrast our work to previous post-calibration methods used in classification and object detection. As the Table suggests, all prior methods use only frequency information and none of them considers the space information.

\section{Methodology}
\label{sec:method}

We show the overview of our approach in Fig.\ref{fig:pipeline}. FRACAL is essentially a post-processing method that calibrates the classification logits of the detector using precomputed weights based on the class and space statistics of the trainset. The FRACAL weights can be stored in the memory, thus during inference our method has insignificant overhead. Its effects on the detector are twofold, on the frequency axis, it decreases frequent class predictions like \textit{hat} and increases rare class predictions like \textit{tiara}. On the space axis, it produces more uniformly spaced predictions for all classes, by forcing e.g. both \textit{hats} and \textit{tiaras} to appear in all locations and not just the top. Next, we analyse our method in detail.

\subsection{Background: Classification Calibration}
\label{subsec:classification_calibration}
Let $f_{y}(x;\theta)=z$ be a classifier parameterised by $\theta$, $x$ the input image, $y$ the class, $z$ the logit, $\bar{y}$ is the model's prediction and $p_s(y)$ and $p_t(y)$ the class priors on the train and test distributions respectively. The post-calibration equation is: 
\begin{equation}
    \begin{split}
    \bar{y}
    =  & \arg \max_y(f_{y}(x;\theta) + \log(p_t(y))
    -\log(p_s(y)) ).
    \end{split}
    \label{eq:label_shift_general}
\end{equation}
This has been numerously analysed in previous literature~\cite{menon2021longtail,alexandridis2023inverse,hong2021disentangling,Ren2020balms,lipton2018detecting} and we derive it in Appendix.
In short, this shows that to get better performance, one can align the model's predictions with the test distribution,  by subtracting $\log(p_s(y))$ and adding $\log(p_t(y))$ in the logit space.
We now extend it to object detection.
\subsection{Classification Calibration for Object Detection}
\label{section:calibration_for_detection}
In classification, $p(y)$ can be easily defined using the dataset's statistics, by using instance frequency $n_y$, 
i.e. $p(y)= \frac{n_y}{\sum_j^C n_j}$. In object detection, this is not the case because $p(y)$ is affected by the location and the object class.
Following \cite{alexandridis2022long}, we define the class priors as:
\begin{equation}
    p(y,o,u) = p(y|o,u) \cdot p(o,u) = p(y,u) \cdot p(o,u),
    \label{eq:class_prior}
\end{equation}
where $o$ is an object, irrespective of class, and $u$ is the location inside the image. 

Substituting Eq.\ref{eq:class_prior} in Eq.\ref{eq:label_shift_general}, $\bar{y}$ becomes:
\begin{equation}
    \begin{split}
    \bar{y}= \arg \max_y(f_{y}(x;\theta) + \log(\frac{p_t(y,u) \cdot p_t(o,u)}{p_s(y,u) \cdot p_s(o,u)} ). \\
    \end{split}
    \label{eq:label_shift_general_obj}
\end{equation}

 The term $p(o,u)$ in Eq.\ref{eq:label_shift_general_obj} cannot be calculated apriori as it depends on the model's training (e.g., the IoU sampling algorithm, how the object class is encoded etc\footnote{Typically object detectors use an extra background logit \textit{bg} to implicitely learn $p(o,u)$.}).
Despite this, $p_s(o,u) \approx p_t(o,u)$, as we show in the Appendix, which means that the object distributions of the train and the test set remain the same and only the foreground class distribution changes.
As a result:
\begin{equation} 
    \bar{y} =\arg \max_y(f_{y}(x;\theta) + \log(p_t(y,u))
    -\log(p_s(y,u))).\\
    \label{eq:label_shift_general2}
\end{equation}
Next, we show how the location parameter $u$ affects Eq.~\ref{eq:label_shift_general2}.
\subsubsection{Location-class independence.}
We consider the case where the location $u$ does not give any information. In this case, $u$ and $y$ are independent variables, thus $p(y,u)=p(y) \cdot p(u)$ and we rewrite Eq.~\ref{eq:label_shift_general2} as:
\begin{equation}
\begin{aligned}
    \bar{y} &=\arg \max_y(f_{y}(x;\theta) + \log(\frac{p_t(y) \cdot p_t(u)}{p_s(y) \cdot p_s(u))})\\
    &=\arg \max_y( f_{y}(x;\theta) + \log(p_t(y))
    -\log(p_s(y))),
    \label{eq:label_shift_general3}
\end{aligned}
\end{equation}
where $p(u)$ is the probability of a random location in the image space and it has been simplified because it is the same in both source and target distributions, i.e., $p_s(u)=p_t(u)$. 
In theory $p_t(y)$ is unknown, thus Eq.\ref{eq:label_shift_general3} cannot be applied. Despite that, we found that setting $p_t(y)=\frac{1}{C}$ works well, because it forces the model to do balanced detections on the test
set.  In practice, this maximises average precision because this metric independently evaluates all classes and it rewards balanced detectors \cite{everingham2010pascal}. Accordingly, the Classification (C) calibration of the logit $z_y$ is:
\begin{equation}
    \text{C}(z_y) =\begin{cases}
            z_y-\log_{\beta}(\frac{n_y}{\sum_i^C n_i})+\log_{\beta}(\frac{1}{C}), \;\; y \in \{1,...,C\}\\ 
            z_y,\;\; y=\text{bg},
        \end{cases}
        \label{eq:ps_softmax_obj}
\end{equation}
where $\beta$ is the base of the logarithm that we optimise through hyperparameter search.  The background logit remains unaffected because of the assumption that the object distribution is the same in train and test set $p_s(o,u) \approx p_t(o,u)$, (this assumption is taken from \cite{pan2021model,zhao2022logit}).
\begin{figure}[!t]
    \centering
    \includegraphics[width=1\linewidth]{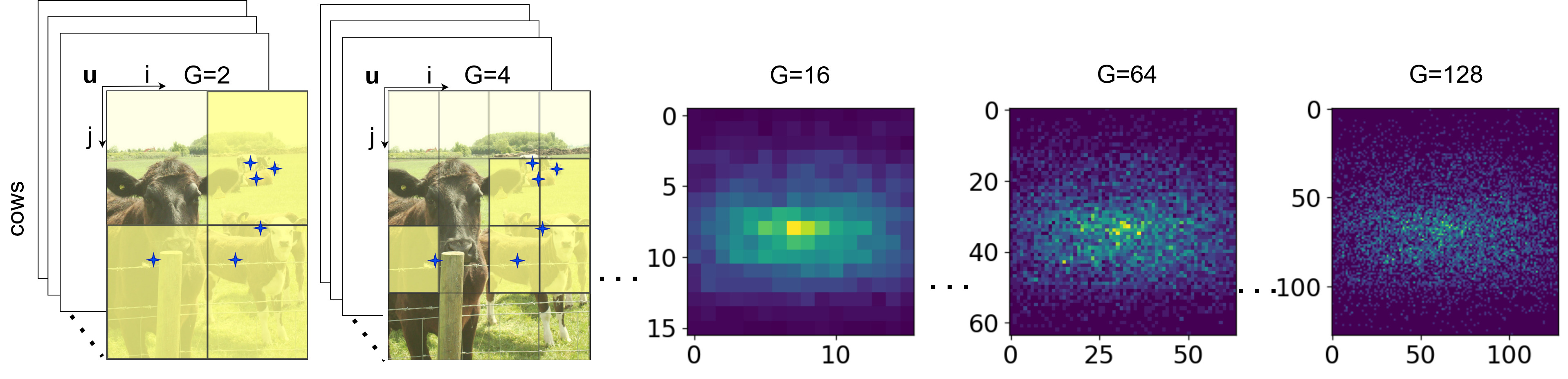}
    \caption{Different grid sizes affect the object distribution estimation. When the grid is coarse, e.g., $1\times1$ or $2\times2$, there is no or little location information. When it is finer, e.g., $64\times64$, the probability is sparse, giving noisy estimates for the rare classes.  }
    \label{fig:grid_sizes}
\end{figure}
To this end, Eq.~\ref{eq:ps_softmax_obj} can get good performance as shown in our ablation study but it is limited because the assumption that $p(y,u)=p(y) \cdot p(u)$ is not correct.
In the real world, the object detection distribution has a strong center bias, as shown in Fig.\ref{fig:grid_sizes} and discussed in \cite{oksuz2020imbalance}. Furthermore, the location is correlated with the class~\cite{kayhan2020translation}, therefore, $p(y,u) \neq p(y) \cdot p(u)$.
As we show, the location provides valuable information for the long-tailed detection task and we enhance Eq.~\ref{eq:ps_softmax_obj} by fusing location information. 

\subsubsection{Location-class dependence.}
One way to compute $p(y,u)$ is by counting the class occurrences $n_{y}(\mathbf{u})$ along locations that fall inside the cell $\mathbf{u}=[i,j]$ as shown in Fig.~\ref{fig:grid_sizes}-left. To do so, we discretise the space of various image resolutions into a normalised square grid $U_{G \times G}$ of fixed size $G \in \mathbf{N}$ and count class occurrences inside every grid cell. Accordingly, the grid dependent calibration is defined as:
\begin{equation}
    \text{C}_{G}(z_{y,\mathbf{u}}) =\begin{cases}
            z_{y,\mathbf{u}}-\log_{\beta}(p_s(y,\mathbf{u}))+\log_{\beta}(p_t(y,\mathbf{u}))\\
            z_{y,\mathbf{u}},\;\;if\;\; y=\text{bg},\\
        \end{cases}
        \label{eq:ps_grid_softmax_obj}
\end{equation}
where $z_{y,\mathbf{u}}$ is the predicted logit whose center falls inside the discrete cell $\mathbf{u}=[i,j]$ and $p_t(y,\mathbf{u})$ is uniform, i.e., $p_t(y,\mathbf{u})=\frac{1}{C} \cdot \frac{1}{G^2}$.

However, the choice of the grid size $G$ largely affects the estimation of $p(y,u)$, as shown in Fig.\ref{fig:grid_sizes}-right. For example, if we use smaller $G$, the generic object distribution becomes denser and little location information is encoded. If we use larger $G$, the distribution becomes sparse. This is problematic for the rare classes because they are already sparse and their location information is noisy. In Table~\ref{tab:ablations}-e, we show that this baseline shows limited performance.

\begin{figure*}[!tb]
    \centering
    \includegraphics[width=0.75\linewidth]{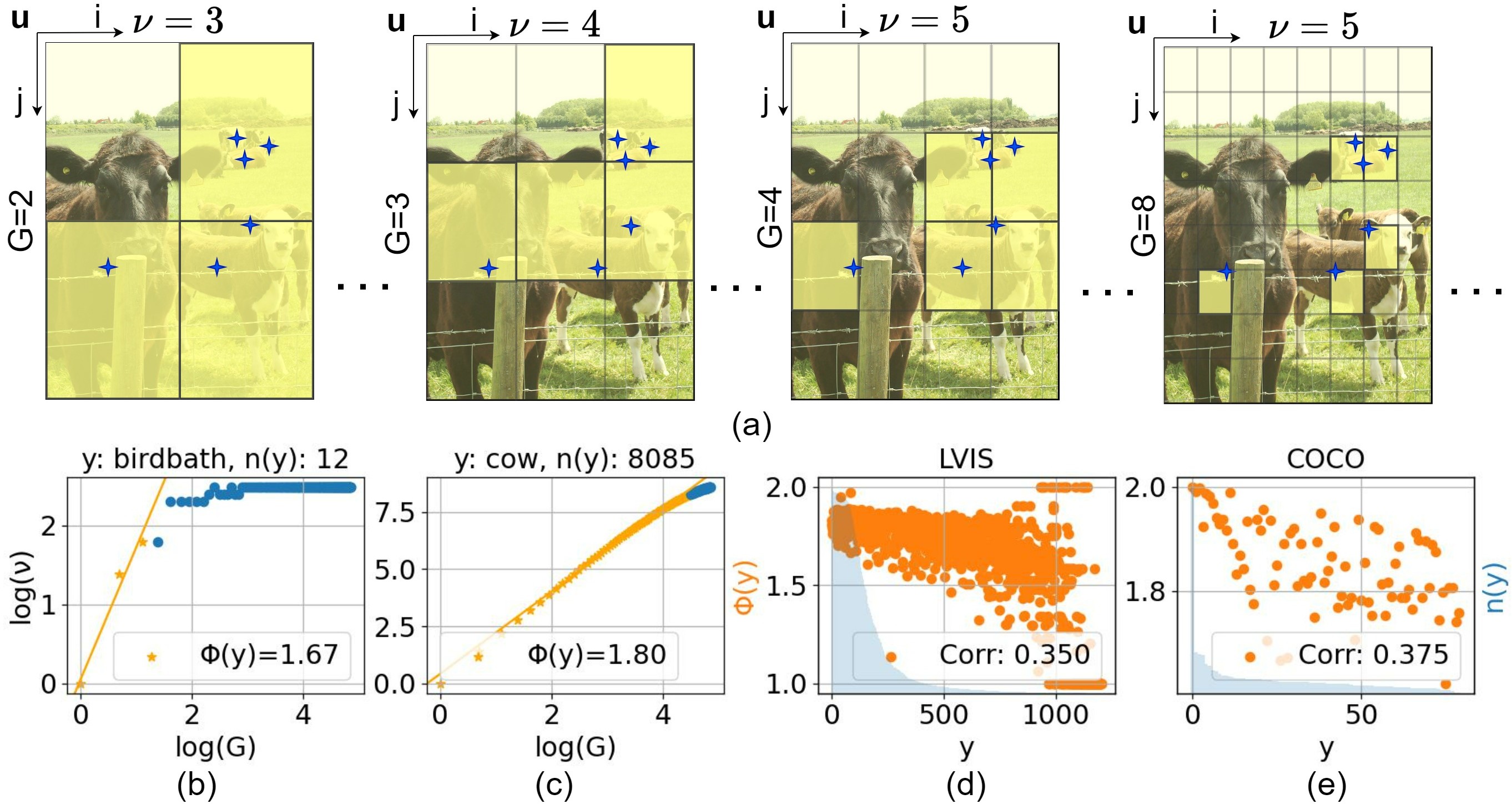}
    \caption{a) An example of the box counting method for the class \textit{cow}. It iteratively counts the boxes $\nu$ containing its center, as $G$ grows. b-c) The blue points are all $G-\nu$ pairs, out of them only the orange points are used to calculate the slope $\Phi$ based on the quadratic rule $G = \lfloor \sqrt{n_y} \rfloor$. d-e) Fractal dimension and class frequency are weakly correlated, showing that the $\Phi$ complements the frequency statistic. }
    \label{fig:fractal_dimension}
\end{figure*}

\subsection{Calibration using fractals}
\label{subsec:location_calibration}
To solve the sparsity problem introduced by the grid-size, we use the fractal dimension $\Phi$~\cite{panigrahy2019differential}, which is a metric independent of the grid size $G$.
To calculate $\Phi$, we use the box-counting method \cite{schroeder2009fractals}:
\begin{equation}
    \Phi(y)= \lim_{G\rightarrow \infty} \frac{\log\sum_{j=0}^{G-1} \sum_{i=0}^{G-1} \mathbbm{1}(n_{y}(\mathbf{u}))}{\log(G)},
    \label{eq:fractal_dimension}
\end{equation}
where $\mathbbm{1}$ is the indicator function. For objects in 2D images, as in our case, $\Phi(y) \in [0,2]$, where 0 is only one object, 1 shows that the objects lie across a line and 2 shows that they are located uniformly across the image space.

For brevity, we rewrite $\nu_y = \sum_{j=0}^{G-1} \sum_{i=0}^{G-1} \mathbbm{1}(n_{y}(\mathbf{u}))$ and we give an example in Fig.~\ref{fig:fractal_dimension}-a.
In practice, Eq.~\ref{eq:fractal_dimension} cannot be computed because by increasing $G$, the computation becomes intractable. Instead, we approximate $\Phi$, by evaluating nominator-denominator pairs of Eq.~\ref{eq:fractal_dimension} for various values of $G$ up to a threshold $t$ and then fit a line to those points. The slope of this line approximates $\Phi(y)$, because it considers all computed $G-\nu_y$ pairs.

\noindent \textbf{Dealing with rare classes.}
To select the threshold $t$, we use the quadratic rule $G\leq t = \lfloor \sqrt{n_y} \rfloor$. The motivation for this rule is simple, for example, if an object is rare, e.g., it appears 4 times in the whole training set, then it can, at most, fill a grid of size $2 \times 2$. For objects with fewer occurrences \textit{we cannot compute $\Phi$ and thus we assign $\Phi=1$}.  Using this rule, we define the maximum number of pairs that are required for fitting the ``fractality'' line highlighted in orange in Fig.~\ref{fig:fractal_dimension}-b and Fig.~\ref{fig:fractal_dimension}-c. For example, the rare object \textit{birdbath} appears 12 times in the training set, thus we use the first three orange points in Fig.~\ref{fig:fractal_dimension}-c that correspond to $G=\{1,2,3\}$, to fit the ``fractality'' line, resulting in a large $\Phi=1.67$. This rule ensures that the fractal dimension computation does not underestimate the rare classes and it gives robust measurements that increase rare class performance as shown in our experiments.
For the \textit{cow} object that has larger frequency we use more $G-\nu$ orange pairs to fit the line as shown in Fig.~\ref{fig:fractal_dimension}-b, resulting in $\Phi=1.80$. 

\noindent \textbf{Relationship to frequency.}
As shown in Fig.~\ref{fig:fractal_dimension}-d and e, the fractal dimension weakly correlates with frequency, i.e., $0.35$ for LVIS and $0.375$ for COCO using Pearson correlation. Also, there are many rare classes with large $\Phi \approx 2$, showing that our threshold selection technique is robust for small sample sets. The weak correlation between frequency and fractal dimension, highlights that our method complements the frequency statistics and adds new information to the model resulting in superior performance as shown experimentally. 

\noindent \textbf{Usage.} After calculating $\Phi$ for all classes in the training set, we store the fractal dimensions in the memory. During inference, we fuse them with the model's prediction using the space-aware calibration (S):
\begin{equation}
    \text{S}(z_y) =
    \begin{cases}
    \frac{\sigma(z_y)}{\Phi(y)^\lambda}, \;\; y \in \{1,...,C\}\\
    \sigma(z_y),\;\; y=\text{bg} ,
    \end{cases}
    \label{eq:location_calibration}
\end{equation}
where $\sigma(z_y) \in (0,1)$ is the model's prediction for class $y$, with $\sigma()$ the Softmax activation, and $\lambda\geq0$ is a hyperparameter. Eq.~\ref{eq:location_calibration} downweighs the classes that appear most uniformly and it upweighs the classes that appear less uniformly. In practice, (S) calibration forces the detector to predict both frequent and rare classes uniformly across all spatial locations. For example, in Fig.~\ref{fig:pipeline}-bottom-right after applying our method, the model predicts \textit{hats} and \textit{tiaras} in every image location and not just the in top of the images as the baseline.  Intuitively, this removes the spatial bias, producing balanced detectors that have better performance as shown in our ablation and our qualitative results. 

\subsection{Localised Calibration}
\label{subsec:local}
By putting Eq.~\ref{eq:ps_softmax_obj} and Eq.~\ref{eq:location_calibration} together, we get the final FRActal CALibration (FRACAL) as:
\begin{equation}
    \text{F}(z_y) = \frac{\text{S}(\text{C}(z_y))}{\sum_{j=1}^{C+1} \text{S}(\text{C}(z_j))} .
    \label{eq:local}
\end{equation}
Our proposed method tackles the classification imbalance using additional space statistics. On the classification axis, we use the class priors $p_s(y)$ and perform logit adjustments.
On the space axis, we use the fractal dimension $\Phi(y)$ to perform a space-aware calibration that accounts for the object's location distribution $p_s(y,u)$. 
In Eq. \ref{eq:local}, we renormalise both foreground and background logits to preserve a probabilistic prediction after the space calibration in Eq. \ref{eq:location_calibration}.

\noindent \textbf{Extending to binary classifiers.}
In long-tailed object detection there are many works that use only binary classifiers \cite{alexandridis2022long,tan2020equalization,tan2021equalization,li2022equalized,wang2021adaptive,hyun2022long,hsieh2021droploss}.
In this case, the logit $z_i$ performs two tasks simultaneously: It discriminates among the foreground classes and performs background-to-foreground classification.  Thus, to correctly apply foreground calibration, we first need to decouple the foreground and background predictions. To do so, we filter out the background proposals using the model's predictions as follows:
\begin{equation}
    \text{F}_{b}(z_i) = \eta(\text{C}(z_i) -\log_{\beta}(\frac{\Phi(y)^\lambda}{\sum_i^C \Phi(i)^\lambda }) +\log_{\beta}(\frac{1}{C})) \cdot \eta(z_i),
        \label{eq:lcd_calib2}
\end{equation}
where $\eta(z_i)$ is the sigmoid activation function that acts as a filter for low-scoring proposals. Compared to Eq.~\ref{eq:local}, Eq. \ref{eq:lcd_calib2} performs class calibration and space calibration in logit space, lowering the false-positive detection rate. 

\vspace{-3mm}
\section{Results}
\label{sec:results}

\begin{table*}[t]
    \centering
    \caption{Comparison against SOTA on LVISv1 dataset. Our method reaches the best results in all metrics.} 
        \vspace*{-1em}
        \begin{tabular}{c|c|c|lllll}
        \toprule
        Method&Reference&Arch. &$AP^m$&$AP^m_r$&$AP^m_c$&$AP^m_f$&$AP^b$ \\
        \hline
         Baseline&N/A&\multirow{7}{*}{Mask RCNN ResNet50}&25.7&15.8&25.1&30.6&25.9\\
         NorCal \cite{pan2021model}&NeurIPS 21&&25.2&19.3&24.2&29.0&26.1\\
         
         GOL \cite{alexandridis2022long}&ECCV 22&&\underline{27.7}&21.4&\underline{27.7}&30.4&27.5\\
         ECM \cite{hyun2022long}&ECCV 22& &27.4&19.7&27.0&\underline{31.1}&27.9\\
         CRAT w/ LOCE~\cite{wang2024learning}&IJCV 24 &&27.5&21.2&26.8&31.0&\underline{28.2}\\
         LogN \cite{zhao2022logit}&IJCV 24&&27.5 &\underline{21.8}&27.1&30.4&28.1\\
         \hdashline\noalign{\vskip 0.3ex}
         \textbf{FRACAL (ours)}&-&&\textbf{28.6}$^\textbf{+0.9}$&\textbf{23.0}$^\textbf{+1.2}$&\textbf{28.0}$^\textbf{+0.3}$&\textbf{31.5}$^\textbf{+0.4}$&\textbf{28.4}$^\textbf{+0.2}$\\
         \hline
         Baseline&N/A&\multirow{8}{*}{Mask RCNN ResNet101}&27.0&16.8&26.5&32.0&27.3\\
         NorCal \cite{pan2021model}&NeurIPS 21&&27.3&20.8&26.5&31.0&28.1\\
         GOL \cite{alexandridis2022long}&ECCV  22&&\underline{29.0}&22.8&29.0&31.7&29.2\\
         ECM \cite{hyun2022long}&ECCV 22& &28.7&21.9&27.9&\underline{32.3}&29.4\\
         ROG \cite{zhang2023reconciling}&ICCV 23&&28.8&21.1&\underline{29.1}&31.8&28.8\\
         CRAT w/ LOCE~\cite{wang2024learning}&IJCV 24 &&28.8&22.0&28.6&32.0&29.7\\
         LogN \cite{zhao2022logit}&IJCV 24&&\underline{29.0} &\underline{22.9}&28.8&31.8&\textbf{29.8}\\\hdashline\noalign{\vskip 0.3ex}
         \textbf{FRACAL (ours)}&-  &&\textbf{29.8}$^\textbf{+0.8}$&\textbf{24.5}$^\textbf{+1.5}$&\textbf{29.3}$^\textbf{+0.2}$&\textbf{32.7}$^\textbf{+0.4}$&\textbf{29.8}\\
         \bottomrule
    \end{tabular}
    \label{tab:sota}
\end{table*}

\subsection{Experimental Setup}
\label{sec:experiment}
We use the Large Vocabulary Instance Segmentation (LVISv1) dataset \cite{gupta2019lvis} which consists of $100k$ images in the train set and $20k$ images in the validation set.
This dataset has $1,203$ classes grouped according to their image frequency into \textit{frequent} (those that contain $>100$ images), \textit{common} (those that contain $10 \sim 100$ images) and \textit{rare} classes (those that contain $<10$ images) in the training set. 
For evaluation, we use average mask precision $AP_{m}$, average box precision $AP_{b}$ and $AP^m_{f}$, $AP^m_{c}$ and $AP^m_{r}$ that correspond to $AP^m$ for \textit{frequent}, \textit{common} and \textit{rare} classes.
Unless mentioned, we use Mask R-CNN \cite{he2017mask} with FPN \cite{lin2017feature}, ResNet50~\cite{he2016deep}, repeat  factor sampler (RFS) \cite{gupta2019lvis}, Normalised Mask \cite{wang2021seesaw}, CARAFE \cite{wang2019carafe} and we train the baseline model using the 2x schedule \cite{he2019rethinking}, SGD, learning rate $0.02$ and weight decay $1e-4$. For Swin models, we train the baseline models with the 1x schedule, RFS, AdamW \cite{kingma2014adam} and $0.001$ learning rate.
During inference, we set the IoU threshold at $0.3$ and the mask threshold at $0.4$. FRACAL is applied before the non-maximum suppression step and it does not have any computational overhead as the weights are precomputed.
\subsection{Main Results}

\noindent \textbf{Comparison to SOTA.} In Table \ref{tab:sota}, we compare FRACAL to the state-of-the-art using ResNet50 and ResNet101.
Using ResNet50, FRACAL significantly surpasses GOL~\cite{alexandridis2022long} by $0.9$ percentage points (pp) in $AP^m$  and by $1.6$pp $AP^m_r$.
On ResNet101 FRACAL achieves $29.8\%$ $AP^m$ and $24.5\%$ $AP^m_r$, outbesting GOL by $0.8\text{pp}$ and $1.7\text{pp}$ respectively. 

FRACAL achieves excellent results not only for rare categories but also for frequent ones, due to the use of fractal dimension, which allows the model to upscale the predictions of frequent but non-uniformly located categories. It achieves $31.5\%$ $AP^m_f$ with ResNet50 and $32.7\%$ $AP^m_f$ with ResNet101 and surpasses the next best method, ECM \cite{hyun2022long} by $0.4\text{pp}$.

Compared to the previous post-calibration method, Norcal \cite{pan2021model}, FRACAL increases performance by $3.4\text{pp}$ $AP^m$, $3.7\text{pp}$ $AP^m_r$, $3.8\text{pp}$ $AP^m_c$, $2.5\text{pp}$ $AP^m_f$ and $2.3\text{pp}$ $AP^b$ using ResNet50. This is because FRACAL boosts both rare and frequent categories via classification and space calibration, respectively, while Norcal only boosts the rare categories and lacks space information.

We also compare our method with Transformer backbones.
Using Swin-T, FRACAL considerably outperforms Seesaw~\cite{wang2021seesaw} by $1.2\text{pp}$ $AP^m$, $1.7\text{pp}$ $AP^m_r$, $1.2\text{pp}$ $AP^m_c$, $1.0\text{pp}$ $AP^m_f$ and $0.8\text{pp}$ $AP^b$ as shown in Table~\ref{tab:sota_swin}. Using Swin-S, FRACAL largely surpasses Seesaw in all metrics and particularly in $AP^m_r$ by $2.2\text{pp}$ which is a significant $8.6\%$ relative improvement for the rare classes.
Finally, we scale our method to Swin-B pretrained on ImageNet22K, and we show that it substantially enhances the $AP^m$ by $1.9\text{pp}$, the $AP^m_r$ by $6.6\text{pp}$ and the $AP^b$ by $2.3\text{pp}$.
\begin{table}[htb]
\centering
\caption{ Results with MaskRCNN, Swin-T/S/B and 1x schedule. }
\vspace*{-1em}
\begin{tabular}{l|lllll}
    \hline
        Method&$AP^m$&$AP^m_r$&$AP^m_c$&$AP^m_f$&$AP^b$ \\ 
        \hline
        Baseline-(T)&27.7&17.9&27.9&31.8&27.1\\
         Seesaw-(T)&\underline{29.5}&\underline{24.0}&29.3&\underline{32.2}&\underline{29.5}\\
         GOL-(T)&28.5&21.1&\underline{29.5}&30.6&28.3\\ \hdashline
         \textbf{FRACAL-(T)}&\textbf{30.7} &\textbf{25.7}&\textbf{30.5}&\textbf{33.2}&\textbf{30.3}\\
         \hline
         Baseline-(S)&30.9&21.7&31.0&34.7&31.0\\
         Seesaw-(S)&\underline{32.4}&\underline{25.6}&\underline{32.8}&\underline{34.9}&\underline{32.9}\\
         GOL-(S)&31.5&24.1&32.3&33.8&32.0\\ \hdashline
         \textbf{FRACAL-(S)}&\textbf{33.6}&\textbf{27.8}&\textbf{33.9}&\textbf{35.9}&\textbf{33.4}\\
         \hline
         Baseline-(B)&36.6&28.9&37.8&\textbf{38.7}&37.1\\
         \textbf{FRACAL-(B)}&\textbf{38.5}&\textbf{35.5}&\textbf{39.4}&\textbf{38.7}&\textbf{39.4}\\
         \bottomrule
    \end{tabular}
\label{tab:sota_swin}
\end{table}

\begin{table}[t]
\centering
\caption{FRACAL can be used with both Sigmoid and Softmax based detectors and improve their precision. }
\vspace*{-1em}
\begin{tabular}{c|cccc}
    \hline
     Method&$AP^b$&$AP^b_r$&$AP^b_c$&$AP^b_f$ \\
     \hline
     ATSS~\cite{zhang2020bridging}&25.3&15.8&23.4&\textbf{31.6}\\
     \textbf{with FRACAL (ours)} &\textbf{26.7}&\textbf{20.8}&\textbf{25.9}&30.9\\
     \hline
     GFLv2~\cite{li2021generalized}&26.6&14.7&25.1&\textbf{33.5}\\
     \textbf{with FRACAL (ours)}&\textbf{28.2}&\textbf{19.4}&\textbf{27.2}&33.2\\
      \hline
     GFLv2 (DCN)~\cite{li2021generalized}&27.4&13.7&26.1&\textbf{34.8}\\
     \textbf{with FRACAL (ours)}&\textbf{28.9}&\textbf{18.7}&\textbf{27.9}&34.5\\
     \hline
     APA \cite{alexandridis2024adaptive}&26.9&14.3&26.2&33.2\\
     \textbf{with FRACAL (ours)}&\textbf{29.2}&\textbf{22.1}&\textbf{28.0}&\textbf{33.7}\\
     \hline
     Cascade RCNN \cite{cai2019cascade}&28.6&16.5&27.8&34.9\\
     \textbf{with FRACAL (ours)}&\textbf{31.5}&\textbf{24.3}&\textbf{31.0}&\textbf{35.3}\\
    \hline
    \end{tabular}
\label{tab:sota_det}
\end{table}

\noindent \textbf{Results on object detectors.}
We evaluate FRACAL with common object detectors in Table \ref{tab:sota_det} using ResNet50. FRACAL boosts the overall and rare category performance of both one-stage detectors such as ATSS~\cite{zhang2020bridging} or GFLv2 \cite{li2021generalized} and two-stage detectors such as Cascade RCNN \cite{cai2019cascade} and APA-MaskRCNN \cite{alexandridis2024adaptive}. Note that on sigmoid-detectors such as ATSS or GFLv2, FRACAL largely boosts the performance of rare and common categories but it slightly reduces the performance of frequent categories. Since the sigmoid activation performs independent classification, the binary version of FRACAL struggles to properly calibrate the predicted unnormalised vector. This limitation was also found in previous works \cite{pan2021model} which also reported that binary logit adjustment produces performance trade-offs between frequent and rare categories. 
For softmax-based detectors, such as Cascade RCNN and APA, FRACAL boosts all categories. In the Appendix, we discuss FRACAL's expected calibration error and detection error

\subsection{Ablation Study and Analysis}
\noindent \textbf{The effect of each module.} FRACAL consists of simple modules that we ablate in Table \ref{tab:ablations}-a. First, MaskRCNN with CARAFE \cite{wang2019carafe}, normalised mask predictor \cite{wang2021seesaw}, cosine classifier \cite{wang2021seesaw} and random sampler achieves $22.8\%$ $AP^m$ and $8.2\%$ rare category $AP^m_r$. On top of this, the fractal dimension calibration (S) improves $AP^m$ and $AP_r^m$ by $2.8\text{pp}$ and $5.5\text{pp}$ respectively.

Using only the classification calibration, (C), $AP^m$ and $AP_r^m$ are enhanced by $3.5\text{pp}$ and $8.3\text{pp}$ respectively, because this technique majorly upweights the rare classes. When (S) is added, then it further increases $AP^m$ by $1.0\text{pp}$ and $AP_r^m$ by $2.5\text{pp}$ compared to only (C), reaching $27.3\%$ $AP^m$ and $19.0\%$ $AP_r^m$. This suggests that (S) is useful and the detector can benefit from space information.  The same trend is observed with RFS in Table \ref{tab:ablations}-d, however, both calibration methods have lower gains because RFS partly balances the classes via oversampling.

\noindent \textbf{Fractal dimension coefficient. }
We ablate the choice of the $\lambda$ coefficient in  Eq.~\ref{eq:location_calibration}. As shown in Table \ref{tab:ablations}-b, the optimal performance is achieved with $\lambda=2$ which increases the rare categories by $0.6\text{pp}$, the common categories by $0.7\text{pp}$, the frequent categories by $0.3$pp, the overall mask performance by $0.6\text{pp}$ and the box performance by $1.0\text{pp}$.

 \noindent \textbf{Class calibration parameter search}. 
 We further ablate the choice of the log base $\beta$ in Eq.~\ref{eq:ps_softmax_obj}, using the most common cases: $2$ (bit), $e$ (nat), and $10$ (hartley).
As shown in Table \ref{tab:ablations}-c, the base-10 is the best as it achieves $26.3\%$ $AP^m$ and $16.5\%$ $AP_r^m$ with the random sampler and $28.0\%$ $AP^m$ and $22.4 \%$ $AP_r^m$ with RFS, thus we use it for all experiments on LVIS.
We also observe that further increasing $\beta$ does not come with a performance improvement.

\begin{table*}[t]
\centering 
\caption{Ablations using MaskRCNN-ResNet50. C and S denote the class and location calibration.}
\vspace*{-1em}
\begin{tabular}{lll}
\resizebox{0.22\linewidth}{!}{%
\begin{tabular}{cc|ccc}
    \hline
         \rotatebox[origin=c]{0}{C}&S&
         $AP^m$&$AP^m_r$\\
         \hline
          &&22.8&8.2\\
         &\checkmark&25.6&13.7\\
         \checkmark&&26.3&16.5\\
         \checkmark&\checkmark&\textbf{27.3}&\textbf{19.0}\\
         \hline
    \end{tabular}%
}
&
\resizebox{0.32\linewidth}{!}{%
\begin{tabular}{c|ccccc}
    \hline
         $\lambda$& $AP^m$&$AP^m_r$&$AP^m_c$&$AP^m_f$&$AP^b$ \\
         \hline
         0.0&28.0&22.4&27.3&31.2&27.4\\
         1.0&28.5&23.0&\textbf{28.0}&\textbf{31.6}&28.3\\ 
         \textbf{2.0}&\textbf{28.6}&23.0&\textbf{28.0}&31.5&\textbf{28.4}\\ 
         3.0&28.5&23.2&\textbf{28.0}&31.5&\textbf{28.4}\\ 
         4.0&28.5&\textbf{23.4}&27.9&31.3&\textbf{28.4}\\ 
    \hline
    \end{tabular}
}&
\resizebox{0.3\linewidth}{!}{
\begin{tabular}{c|cc|cc}
    \hline
         \multirow{2}{*}{$\beta$} &\multicolumn{2}{c|}{random}&\multicolumn{2}{c}{RFS}\\
         &$AP^m$&$AP^m_r$&$AP^m$&$AP^m_r$\\
         \hline
            2&19.9&14.7&19.9&18.8\\
            $e$&25.1&16.6&25.8&21.1\\
            \textbf{10}&26.3&16.5&\textbf{28.0}&\textbf{22.4}\\
    \hline
    \end{tabular}
}
\\
\specialcell[t]{\small(a) Results with random sampler.}
&\specialcell[t]{\small(b) Ablation of $\lambda$, using RFS.}
&\specialcell[t]{\small(c) Ablation of $\beta$, under various samplers.}\\
\resizebox{0.22\linewidth}{!}{%
\begin{tabular}{cc|ccc}
    \hline
         \rotatebox[origin=c]{0}{C}&S&
         $AP^m$&$AP^m_r$\\
         \hline
         & &25.7&15.8\\
        &\checkmark&27.7&20.7\\
        \checkmark&&28.0&22.4\\
        \checkmark&\checkmark&\textbf{28.6}&\textbf{23.0}\\
        \hline
    \end{tabular}%
}&
\resizebox{0.32\linewidth}{!}{%
    \begin{tabular}{c|ccccc}
    \toprule
         Method& $AP^m$&$AP^m_r$&$AP^m_c$&$AP^m_f$&$AP^b$  \\
         \midrule
         G=1&28.0&22.4&27.3&31.2&27.4\\
         G=2&27.1&17.5&27.2&31.1&26.6\\
         G=4&25.0&10.5&25.4&31.1&24.9\\
         \hline
         \textbf{ours}&\textbf{28.6}&\textbf{23.0}&\textbf{28.0}&\textbf{31.5}&\textbf{28.4}\\
    \bottomrule
    \end{tabular}
}&
\resizebox{0.34\linewidth}{!}{
\begin{tabular}{c|c|c|c}
    \toprule
         Method&$AP^m$&$AP^m_r$&$AP^b$  \\
         \hline
         FRACAL-Opposite&27.4 &20.5 &26.9 \\ 
         \textbf{FRACAL} &\textbf{28.6}&\textbf{23.0}&\textbf{28.4}\\ 
    \bottomrule
    \end{tabular}
}
\\\specialcell[b]{\small(d) Results using RFS \cite{gupta2019lvis}.}
&\specialcell[b]{\small(e) Comparison against Grid-based methods.}
&\specialcell[b]{\small(f) FRACAL fusion ablation study. }
\\
\end{tabular}

\label{tab:ablations}
\end{table*}


\noindent \textbf{Comparison to grid-dependent calibration.}
We compare FRACAL against the grid-based method, Eq.~\ref{eq:ps_grid_softmax_obj},  in Table \ref{tab:ablations}-e.
When $G=1$ the method does not consider any location information because all predictions fall inside the same grid cell. This achieves the second best performance and it is the same result with the $\lambda=0$ of Table \ref{tab:ablations}-b. When the grid size $G$ is enlarged, the performance of the rare classes drops significantly because the estimated prior distribution $p_s(y,\mathbf{u})$ becomes sparse (see Fig. \ref{fig:grid_sizes}).
FRACAL does not suffer from this problem, because it re-weights all classes based on fractal dimension.

 \noindent \textbf{FRACAL Opposite. } We further test, FRACAL-Opposite which is a variant that applies an invert weighting logic, (i.e. it upweights the uniformly located classes and downweights the non-uniform ones, which rectifies the space bias). As Table \ref{tab:ablations}-f shows, our standard FRACAL achieves better $AP^m$ and $AP^b$ than the Opposite. This shows that it is preferable to remove spatial bias from the object detectors rather than rectify it, because it leads to balanced detectors.

\noindent \textbf{Generalisation to other datasets.}
We test FRACAL on MS-COCO~\cite{lin2014microsoft}, V3DET ~\cite{wang2023v3det} and OpenImages~\cite{kuznetsova2020open} to understand its generalisation ability and report the results in Tables \ref{tab:coco_results},\ref{tab:v3det},\ref{tab:openimages} respectively. The first two datasets are fairly balanced therefore, we do not expect our long-tailed designed detector to massively outperform the others. In Table \ref{tab:coco_results}, FRACAL increases the performance of all models, by an average of $0.5\text{pp}$ $AP^b$ and $AP^m$. In Table \ref{tab:v3det}, FRACAL increases the performance of APA \cite{alexandridis2024adaptive} by $0.4\text{pp}$ $AP^b$. In Table \ref{tab:openimages}, we show that FRACAL outperforms ECM using CascadeRCNN by $1.7\text{pp}$ and it further increases the performance of CAS by $2.0\text{pp}$ and $1.2\text{pp}$ using FasterRCNN and CascadeRCNN respectively.

\begin{figure}
    \centering
    \includegraphics[width=1.0\linewidth]{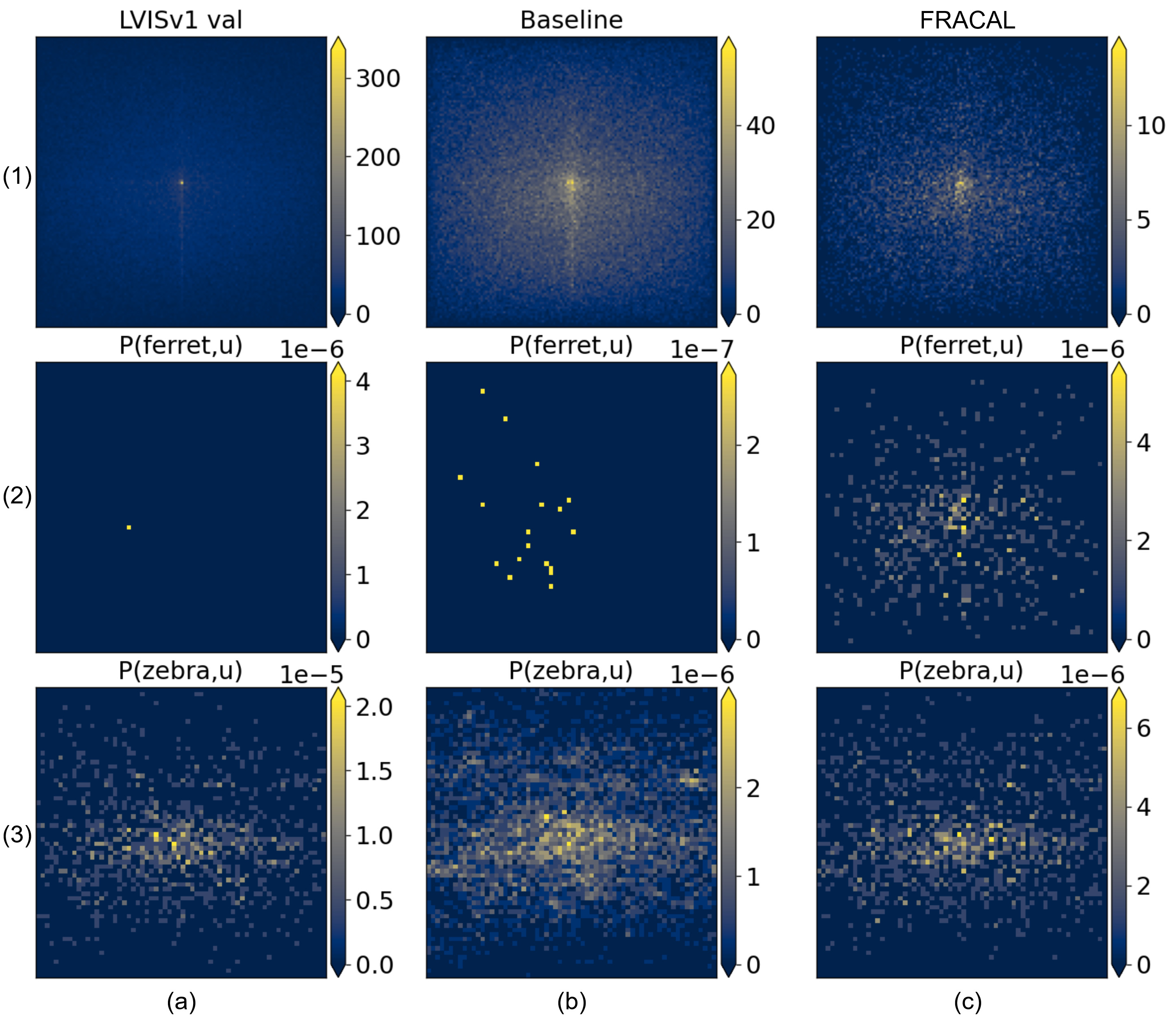}
    \caption{Detection results in LVIS, FRACAL detects more uniformly in both frequency and space axis compared to the baseline.}
    \label{fig:local_preds}
\end{figure}

\noindent \textbf{Qualitative Analysis.}
In Fig. \ref{fig:local_preds}, we show: (a) the ground truth distribution, (b) the baseline and (c) FRACAL predicted distributions concerning all objects (1), the rare class \textit{ferret} (2) and the frequent class \textit{zebra} (3).
FRACAL achieves better precision than the baseline because it detects more rare objects in (2-c) increasing recall and fewer frequent objects in (3-c) decreasing false positives. Regarding the spatial distributions, FRACAL increases the spatial uniformity of all predictions because it has less centered detections for all objects in (1-c), more evenly spread detections for the \textit{ferret} in (2-c) and less centrally biased detections for the \textit{zebra} in (3-c).  This shows that FRACAL makes balanced predictions in both frequency and space perspectives, enabling higher detection performance.

\begin{table}[htb]
    \centering
    \caption{Results on COCO with MaskRCNN.} 
    \begin{tabular}{c|cc|cc}
    \hline
         Method&$AP^m$&$AP^m_{50}$&$AP^b$&$AP^b_{50}$\\
         \hline
         ResNet-50 \cite{he2016deep}&35.4&56.7&39.4&59.9\\
         with FRACAL &\textbf{35.8}&\textbf{57.5}&\textbf{39.9}&\textbf{60.6}\\
        \hline
         SE-ResNet-50 \cite{hu2018squeeze}&36.9&58.8&40.5&61.7\\
         with FRACAL &\textbf{37.4}&\textbf{59.5}&\textbf{41.1}&\textbf{62.4}\\
        \hline
        CB-ResNet-50 \cite{woo2018cbam} &37.3&59.2&40.9&62.1\\
         with FRACAL &\textbf{37.8}&\textbf{60.2}&\textbf{41.5}&\textbf{62.9}\\
        \hline
        Swin-T \cite{liu2021swin}&41.6&65.3&46.0&68.2\\
        with FRACAL &\textbf{41.9}&\textbf{66.0}&\textbf{46.4}&\textbf{68.7}\\
        \hline
    \end{tabular}
    \label{tab:coco_results}
\end{table}

\begin{table}[t]
\caption {Results on V3Det~\cite{wang2023v3det} using FasterRCNN ResNet50.}
\centering
\begin{tabular}{c|c|c|c}
    \hline
         Method&$AP^b$&$AP^b_{50}$&$AP^b_{75}$\\
         \hline
         Normalised Layer \cite{wang2021seesaw}&25.3&32.8&28.1\\
         APA \cite{alexandridis2024adaptive} &29.9&37.6&32.9\\
         APA + FRACAL  (ours) &\textbf{30.3}&\textbf{37.7}&\textbf{33.2}\\
        \hline
    \end{tabular}
    \label{tab:v3det}
\end{table}

\begin{table}[t]
\caption { Results on OpenImages \cite{kuznetsova2020open} using ResNet50.}
\centering
\begin{tabular}{c|c|c}
    \hline
         Method&Detector&$AP^b_{50}$\\
        \hline
        CAS~\cite{liu20201st}&\multirow{2}{*}{Faster RCNN}&65.0\\
        CAS + FRACAL (ours)&&\textbf{67.0}\\
        \hline
        ECM~\cite{hyun2022long}&\multirow{3}{*}{Cascade RCNN}&65.8\\
        CAS~\cite{liu20201st}&&66.3\\
        CAS + FRACAL  (ours)&&\textbf{67.5}\\
        \hline
    \end{tabular}
    \label{tab:openimages}
\end{table}

\section{Conclusion}
\label{sec:conclusion}
We propose FRACAL, a novel post-calibration method for long-tailed object detection.
Our method performs a space-aware logit adjustment, utilising the fractal dimension and incorporating space information during calibration. FRACAL majorly boosts the performance of the detectors by detecting rare classes that are evenly spread inside the image. We show that FRACAL can be easily combined with both one-stage Sigmoid detectors and two-stage Softmax segmentation models.
Our method boosts the performance of detectors by up to $8.6\%$ without training or additional inference cost, surpassing the SOTA in the LVIS benchmark and generalising well to COCO, V3Det and OpenImages.

\noindent \textbf{Acknowledgments.} Anh Nguyen was supported by Royal Society ISPF International Collaboration Awards 2023 (Japan) ICA\textbackslash R1\textbackslash 231067. The project was co-funded by ViTac: Visual-Tactile Synergy for Handling Flexible Materials" (EP/T033517/2).

{
    \small
    \bibliographystyle{ieeenat_fullname}
    \bibliography{main}
}

\clearpage
\maketitlesupplementary

\section{Background: Classification Calibration}
\label{section:classification_calibration_supp}
We theoretically derive the classification calibration for image classification.
Let $p_s(y|x)$  and $p_t(y|x)$ be the source and target conditional distributions. Using the Bayes theorem, we write the source and target conditional distributions as:
\begin{equation}
    p_s(y|x) = \frac{p_s(x|y)p_s(y)}{p_s(x)} \;, p_t(y|x) = \frac{p_t(x|y)p_t(y)}{p_t(x)}
    \label{eq:p_source_cond}
\end{equation}
Dividing them, we write the target conditional distribution:
\begin{equation}
    \begin{split}
        p_t(y|x) = & \frac{1}{\kappa(x)}\frac{p_t(y)}{p_s(y)} p_s(y|x)\frac{p_t(x|y)}{p_s(x|y)}
    \label{eq:dist_shift}
    \end{split}
 \end{equation}
where $\kappa(x)=\frac{p_t(x)}{p_s(x)}$.
During training, we approximate $p_s(y|x)$ by model $f_{y}(x;\theta)=z$ and a scorer function $s(x)=e^x$ for multiple category classification. Thus, the learned source conditional distribution is $p_s(y|x) \propto e^{f_{y}(x;\theta)}$.
 Substituting it inside Eq. \ref{eq:dist_shift}, we rewrite the target condition distribution as:
\begin{equation}
\begin{split}
    p_t(y|x) \propto &  \frac{1}{\kappa(x)}\frac{p_t(y)}{p_s(y)} e^{f_{y}(x;\theta)}\frac{p_t(x|y)}{p_s(x|y)}\\
     = & d(x,y) \cdot e^{f_{y}(x;\theta) + \log(p_t(y)) - \log(p_s(y)) - \log(\kappa(x))} \raisetag{2.5\baselineskip}
    \label{eq:pc_softmax}
\end{split}
\end{equation}
where we assume that $d(x,y)=\frac{p_t(x|y)}{p_s(x|y)}=1$. This is a reasonable assumption, in cases where both train and test generating functions come from the same dataset, as it is in our benchmarks. In inference, we calculate the prediction $\bar{y}$ by taking the maximum value of Eq. \ref{eq:pc_softmax}:
\begin{equation}
    \begin{split}
    \bar{y}
    =& \arg \max_y e^{ (f_{y}(x;\theta) + \log(p_t(y))
    -\log(p_s(y)) - \log(\kappa(x))) }\\
    =  & \arg \max_y(f_{y}(x;\theta) + \log(p_t(y))
    -\log(p_s(y)) )
    \end{split}
    \label{eq:label_shift_general_supp}
\end{equation}
where $\kappa(x)$ is simplified because it is a function of $x$ and it is invariant to $\arg \max_y$.
Eq. \ref{eq:label_shift_general_supp} is the post-calibration method \cite{menon2021longtail,hong2021disentangling}. It can be used during inference to achieve balanced performance by injecting prior knowledge inside the model's predictions, via $p_t(y)$ and $p_s(y)$, in order to align the source with the target label distribution and compensate for the label shift problem. 

\section{Fractal Dimension Variants}
\label{sec:fractal_dim_variants}
 We explore various ways for computing the fractal dimension using the box-counting method \cite{schroeder2009fractals}, the information dimension \cite{renyi1959dimension} (Info), and a smooth variant (SmoothInfo).
 The information variant is defined as:
\begin{equation}
    \text{Info-}\Phi(y)= \lim_{G\rightarrow \infty} \frac{\log\sum_{j=0}^{G-1} \sum_{i=0}^{G-1} \frac{\mathbbm{1}(n_{y}(\mathbf{u}))}{G^2}}{\log(G)}
    \label{eq:info_fractal_dimension}
\end{equation}
It is the similar to the box-counting dimension, except for the box count which is normalised by the grid size $G^2$. This way, the information dimension is represented by the growth rate of the probability $p = \frac{\mathbbm{1}(n_{y}(\mathbf{u}))}{G^2}$ as $G$ grows to infinity.

In practise, the quantity $\mathbbm{1}(n_{y}(\mathbf{u}))$ can be frequently zero for many locations $\mathbf{u}$ especially for rare classes that have few samples and are sparsely located. For this reason, we also proposed a smooth information variant defined as:
\begin{equation}
    \text{Smooth-}\Phi(y)= \lim_{G\rightarrow \infty} \frac{1+\log\sum_{j=0}^{G-1} \sum_{i=0}^{G-1} \frac{1+\mathbbm{1}(n_{y}(\mathbf{u}))}{G^2}}{\log(G)}
    \label{eq:smooth_info_fractal_dimension}
\end{equation}
This Equation is inspired by the smooth Inverse Document Frequency \cite{robertson2004understanding} used in natural language processing and its purpose is to smooth out zero values in $\mathbbm{1}(n_{y}(\mathbf{u}))$ calculation.

\begin{table}[t]
    \centering
    \begin{tabular}{c|ccc}
    \hline
         Dimension& $AP^m$&$AP^m_r$&$AP^b$ \\
         \hline
         Info&\textbf{28.6}&23.2&28.3\\  
         SmoothInfo&\textbf{28.6}&\textbf{23.4}&28.3\\  
         \textbf{Box}&\textbf{28.6}&23.0&\textbf{28.4}\\  
    \hline
    \end{tabular}
    \caption{Fractal Dimension Variants using MaskRCNN with ResNet50 and RFS on LVISv1. All of the are robust and we have chosen the Box variant in the main paper.}
    \label{tab:dimension_variants}
\end{table}

 All variants are robust and SmoothInfo achieves slightly better $AP^m_r$ because its calculation is more tolerant to few samples compared to the box-counting method. However, SmoothInfo and Info achieve slightly worse $AP^b$, thus we use the box-counting method in the main paper.

\begin{table*}[t]
    \centering
    \begin{tabular}{c|cccccc}
    \hline
         Method& $dAP^b_{Cls}$($\downarrow$)&$dAP^b_{Loc}$($\downarrow$)&$dAP^b_{Both}$($\downarrow$)&$dAP^b_{Dupe}$($\downarrow$)&$dAP^b_{Bkg}$($\downarrow$)&$dAP^b_{Miss}$($\downarrow$)\\
         \hline
         Baseline&31.76&\textbf{6.16}&\textbf{0.45}&0.32&\textbf{1.8}&\textbf{6.82}\\
         Cls calibration&20.49&6.96&1.02&\textbf{0.01}&3.26&6.46\\
         Cls + Space calibration&\textbf{16.91}&6.42&0.85&\textbf{0.01}&2.84&6.84\\
        \hline
    \end{tabular}
    \caption{Error analysis using TIDE toolkit \cite{bolya2020tide}. The class calibration reduces the misclassification error but it introduces more false background detections compared to the baseline. When adding the space calibration, it further reduces the misclassification error and also reduces the false background detections compared to the class calibration only.}
    \label{tab:error_analysis}
\end{table*}

\section{Object Distributions}
\begin{figure}[htb]
    \centering
    \includegraphics[width=0.8\linewidth]{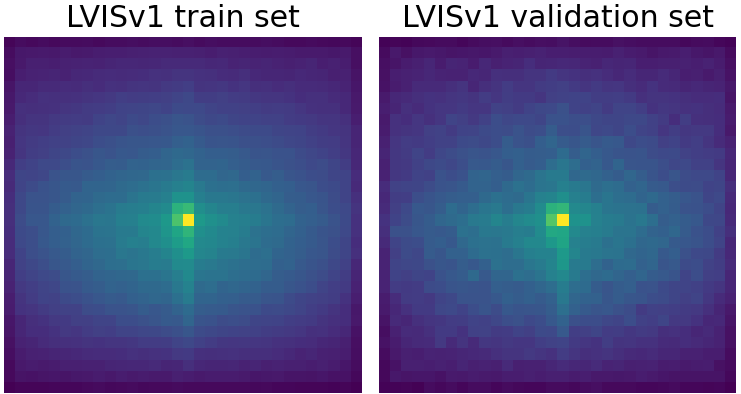}
    \caption{Comparison between the $p_s(o,u)$ (left) and $p_t(o,u)$ (right) in LVISv1 dataset. The distributions are similar, therefore we can safely assume that $p_s(o,u) \approx p_t(o,u)$. }
    \label{fig:lvis_trainval}
\end{figure}

We show that the object distribution $p_s(o,u)$ in the training set is similar to the object distribution $p_t(o,u)$ on the test set in the LVIS v1 dataset \cite{gupta2019lvis}. As shown in Figure \ref{fig:lvis_trainval}, the distributions are close therefore we can safely assume that  $p_s(o,u) \approx p_t(o,u)$. This explains the reason why the background logit should remain intact during calibration because there does not exist label shift for the generic object class (also for the background class) between the train and test sets.

\section{Error Analysis}
We perform an error analysis on the baseline method that uses MaskRCNN ResNet50 and our FRACAL with the same architecture, using the TIDE toolkit \cite{bolya2020tide}. The TIDE toolkit, reports errors using the $dAP^b$ metric which shows the $AP^b$ loss due to a 
specific detection error. In more detail, $dAP^b_{Cls}$ shows the $AP^b$ loss due to mis-classification, $dAP^b_{Loc}$ shows the $AP^b$ loss due to mislocalisation, $dAP^b_{Both}$ indicates the loss due to both misclassification and mislocalisation, $dAP^b_{Dupe}$ indicates the loss due to duplicate detections, $dAP^b_{Bkg}$ shows the error due to background detections and $dAP^b_{Miss}$ shows the errors due to miss-detections. 

As Table \ref{tab:error_analysis} shows, by adding only the class calibration, the $dAP^b_{Cls}$ is reduced significantly by $11.29$ percentage points (pp) compared to the baseline. This shows that the class calibration method increases the correct rare class predictions. However, the class calibration sometimes overestimates the rare classes and predicts rare objects instead of background obtaining $1.44$pp larger $dAP^b_{Bkg}$ than the baseline. When the space calibration is added to the class calibration then it further reduces the misclassification rate $dAP^b_{Cls}$ by $3.58$pp showing that the space calibration classifies better the rare classes. At the same time, the space calibration slightly reduces the false background predictions by $0.42$pp compared to the class calibration method. This analysis shows that both class and space calibration are important and the joint use can lead to better and more balanced detectors.

\section{Confidence Calibration.}
Many works in classification \cite{kull2017beta,platt1999probabilistic,guo2017calibration,niculescu2005predicting} and object detection \cite{kuppers2020multivariate,pathiraja2023multiclass, kuzucu2025calibration} study confidence calibration, which is a technique that allows the model to match its confidence score with its expected accuracy. Confidence calibration is important because it allows the detectors to output calibrated predictions that match the expected average precision. This leads to a safer deployment of detectors because the calibrated detectors provide reassurance regarding their detections which is a desired property in many safe-critical applications like autonomous vehicles \cite{kuzucu2025calibration}. 

In our main paper, we have focused on logit calibration, following the terminology of \cite{pan2021model}, which is different from confidence calibration, as the former aims in improving the rare class performance and the latter aims in reducing the expected calibration error. Despite that, we have analysed the calibration performance of our method compared to the baseline using the LVIS validation set, MaskRCNN ResNet50 and the newly proposed $LaECE_0$ and $LaACE_0$ metrics \cite{kuzucu2025calibration}. The $LaECE_0$ metric shows the mean absolute error between the detection confidence of all predictions and the respective IoU that matches the ground-truth aggregated over 25 confidence bins. $LaACE_0$ is similar to $LaECE_0$, however instead of using 25 bins, it uses an adaptive bin size. Both metrics jointly measure the localisation and classification quality and a low value indicates that the detection has an aligned localisation and classification estimate for the ground-truth. For more details on these metrics, please refer to \cite{kuzucu2025calibration}.  
\begin{table}[htb]
    \centering
    \begin{tabular}{c|c|c}
    \hline
         Method&$LaECE_0$ ($\downarrow$) & $LaACE_0$ ($\downarrow$) \\
         \hline
         Baseline&16.8& 19.8\\
         FRACAL (ours) &\textbf{14.9}& \textbf{15.1}\\
    \hline
    \end{tabular}
    \caption{Calibration Errors using MaskRCNN ResNet50 on LVIS validation set. FRACAL reduces the calibration error compared to the baseline.}
    \label{tab:calibration_error}
\end{table}

As Table \ref{tab:calibration_error} shows, FRACAL reduces the calibration error by $1.9$ $LaECE_0$ points
and by $4.7$ $LaACE_0$ points, compared to the baseline. This shows that FRACAL logit calibration does not produce unreasonable class confidence estimates, in contrast, it enhances the confidence estimates for the rare classes making FRACAL a suitable logit adjustment method for long-tailed object detection. 

\section{Visualisations}

\subsection{Visualisation of the $\Phi$ distribution}
\begin{figure}[htb]
    \centering
    \includegraphics[width=1\linewidth]{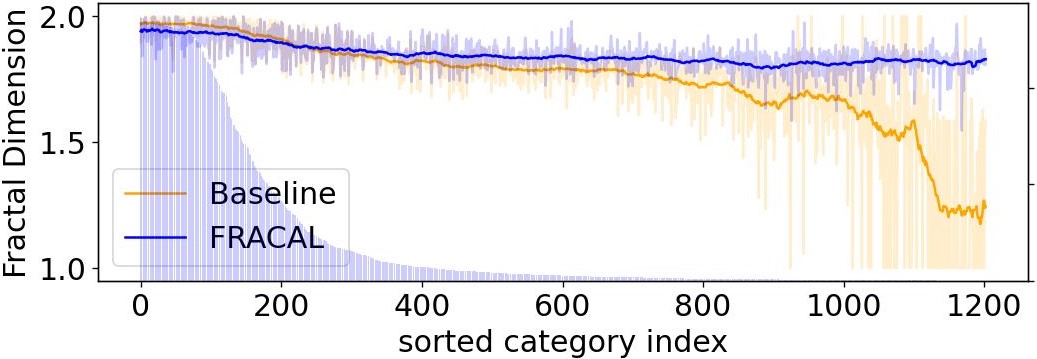}
    \caption{$\Phi$ distributions of the detected objects.}
    \label{fig:phi_distribution}
\end{figure}

As shown in the Figure \ref{fig:phi_distribution}, after applying our method, the detected objects have larger $\Phi$ values than the baseline, especially for the rare classes, which means that the detector makes more spatially balanced detections than the baseline.
\subsection{Detections using FRACAL}
\begin{figure*}[htb]
    \centering
    \includegraphics[width=1\linewidth]{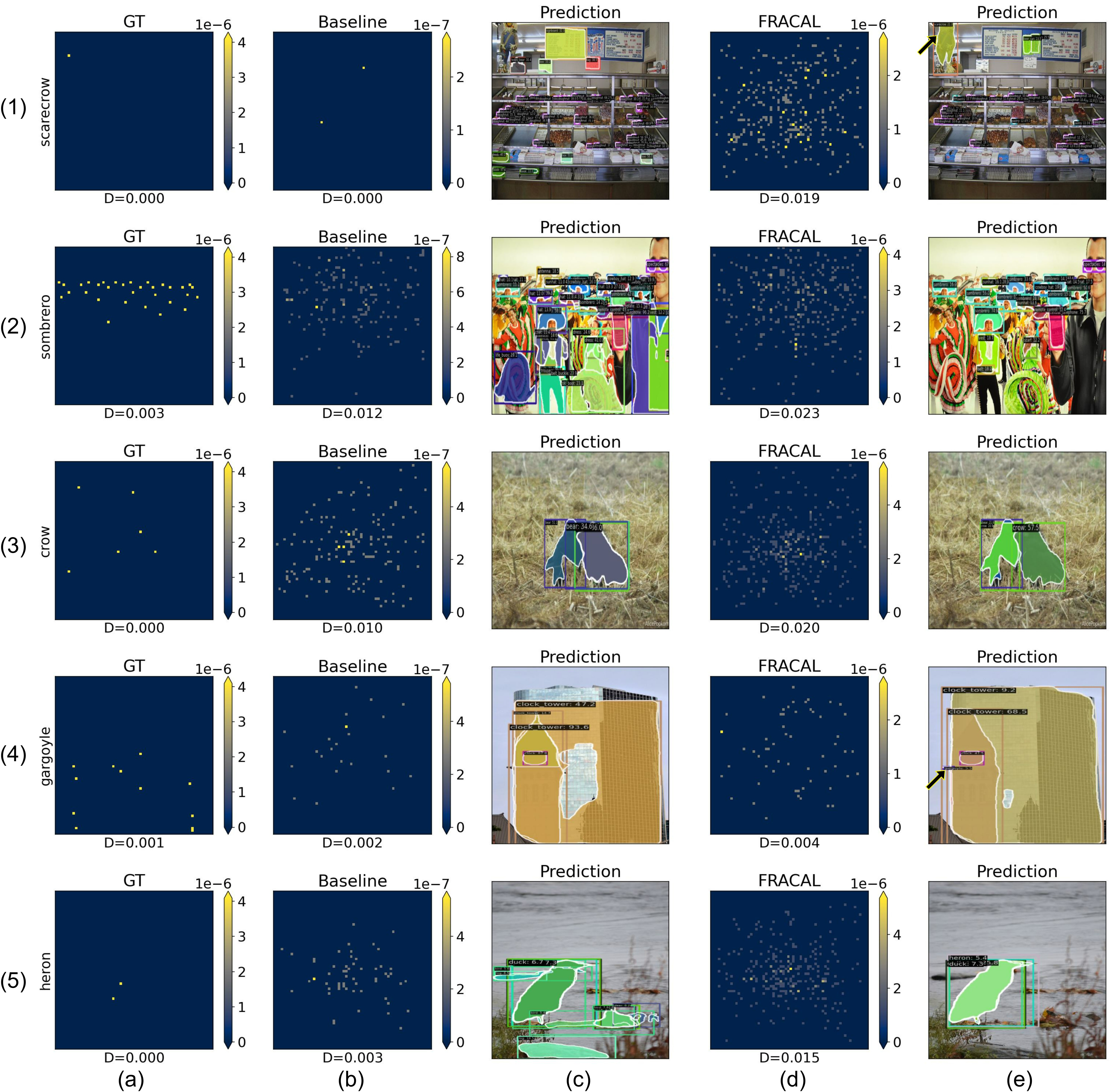}
    \caption{Visualisations on LVIS validation set using MaskRCNN ResNet 50 and mmdetection framework. It is recommended to zoom in for better visualisation of the detections. Our FRACAL, makes more spatially balanced predictions indicated by the larger distance $D$,  and it detects more rare classes compared to the baseline.}
    \label{fig:fracal_vis}
\end{figure*}

 We provide more visualisations of our method in Figure \ref{fig:fracal_vis}. In (a) the ground-truth $p(y,u)$ of LVIS validation set is displayed using a grid of 64 by 64. In (b) the Baseline $\tilde{p}_b(y,u)$ is shown, with an one example prediction in (c) using a detection threshold of $0.01$. In (d) our FRACAL $\tilde{p}_f(y,u)$ is shown using the same detection threshold and finally in (e) one example prediction of FRACAL is shown. In these subfigures, the name of the y-axis denotes the class name and the name of the x-axis indicates the average euclidean distance $D$ between the coordinates of the predictions and the center point in image space. This distance measure shows how spread-out are the predictions and a larger distance indicates larger spatial uniformity because more predictions are further away from the image center.
 As the Figure suggests, our FRACAL predictions in (d) have larger distance $D$ from the image center compared to the baseline predictions in (b), highlighting that FRACAL is more spatially balanced than the baseline. Furthermore, FRACAL makes more rare class-predictions as shown in (d) than the baseline predictions shown in (b), and correctly retrieves the rare classes \textit{scarecrow}, \textit{sobrero}, \textit{crow}, \textit{gargoyle} and \textit{heron} in the (1-e), (2-e), (3-e), (4-e) and (5-e) respectively, in contrast to the baseline that fails to detect them.

\section{FRACAL computational cost}
FRACAL needs 28 seconds to compute the fractal dimension of all objects in LVIS dataset, using multiprocessing with 24 CPU cores. In comparison, LVIS inference costs 90 minutes, using 4-V100, thus our method only accounts for $0.5\%$ of the total computation.

\end{document}